\documentclass{article}
\usepackage{spconf,amsmath,graphicx,tikz}
\usepackage{amsfonts}
\usepackage{kotex}
\usepackage{algorithm}
\usepackage{algorithmic}
\usepackage{newfloat}
\usepackage{listings}
\usepackage{todonotes}
\usepackage{color, colortbl}
\usepackage{url}

\def\ea{\emph{et al.}\,}
\def\ni{\noindent}

\definecolor{LightCyan}{rgb}{0.88,1,1}

\title{LEAP:D - A Novel Prompt-based Approach for Domain-Generalized Aerial Object Detection}

\name{Chanyeong Park\textsuperscript{\rm $\dagger$}, Heegwang Kim\textsuperscript{\rm $\dagger$}, and Joonki Paik\textsuperscript{\rm *} \thanks{ $\dagger$ Co-first authors: chanyeong@ipis.cau.ac.kr and heegwang@ipis.cau.ac.kr.  * Corresponding author: paikj@cau.ac.kr. \, This work was supported by Institute of Information \& communications Technology Planning \& Evaluation (IITP) grant funded by the Korea government(MSIT) (2021-0-01341,  Artificial Intelligence Graduate School Program(Chung-Ang University)), and National Research Foundation of Korea(NRF) grant funded by the Korea government(MSIT)(NRF-RS2024-00343863).}}
\address{Department of Image, Chung-Ang University
  }

\begin{document}
%
\maketitle
\begin{abstract}
Drone-captured images present significant challenges in object detection due to varying shooting conditions, which can alter object appearance and shape. 
Factors such as drone altitude, angle, and weather cause these variations, influencing the performance of object detection algorithms. 
To tackle these challenges, we introduce an innovative vision-language approach using learnable prompts. 
This shift from conventional manual prompts aims to reduce domain-specific knowledge interference, ultimately improving object detection capabilities. 
Furthermore, we streamline the training process with a one-step approach, updating the learnable prompt concurrently with model training, enhancing efficiency without compromising performance. 
Our study contributes to domain-generalized object detection by leveraging learnable prompts and optimizing training processes. 
This enhances model robustness and adaptability across diverse environments, leading to more effective aerial object detection.
\end{abstract}
\begin{keywords}
Object Detection, Domain Generalization, Vision-Language Model
\end{keywords}

\begin{figure*}[!t]
\centering
\includegraphics[width=7in]{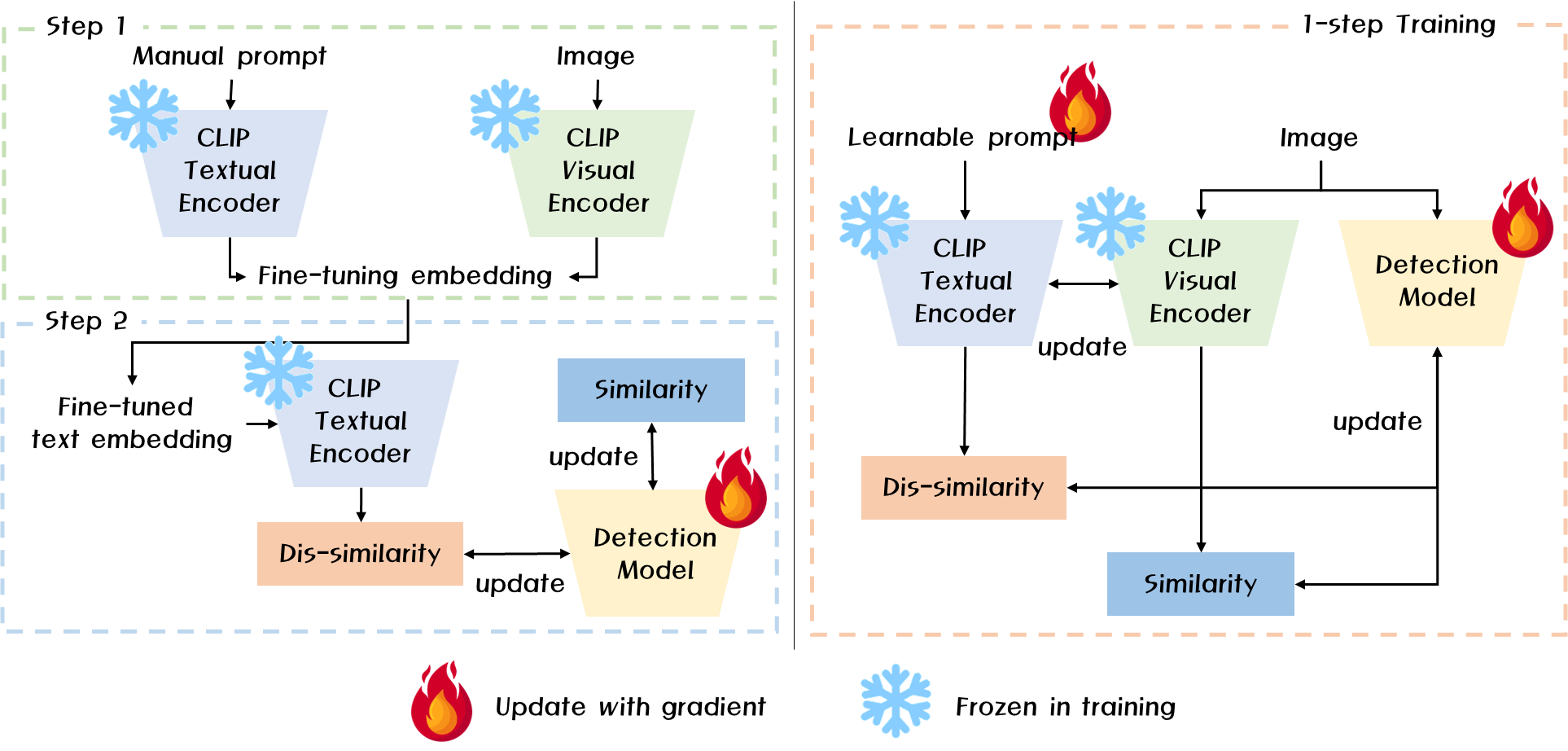}%
\caption{Comparison between the traditional two-step training strategy (left) and the one-step training strategy of the proposed LEAP:D (right). In step 1, text embeddings are fine-tuned to align with the training domain through a two-step training process. In contrast, the proposed method (right) uses learnable embeddings, allowing training without specific alignment to the target domain, thus facilitating a streamlined one-step process.
}
\label{fig.proposed}
\end{figure*}

\section{Introduction}
\label{sec:intro}
Object detection in aerial imagery is rapidly progressing in tandem with advancements in deep learning.
However, the unique characteristics of drone imagery pose significant challenges, impacting object appearance and shape due to varying shooting conditions. 
As a result, there is considerable interest in employing domain generalization techniques to address these challenges. 
Factors such as drone altitude, angle, and weather conditions contribute to variability in shooting conditions. 
Variations in drone altitude can affect object sizes, while camera viewing direction can alter object shapes. 
Moreover, weather conditions and the time of day when images are captured cause significant differences in lighting, further influencing the imagery.

To simply enhance the object detection performance in drone captured images, \cite{kim2023high, park2024enhanced} conducted research on methods for processing high-resolution images. The Nuisance Disentangled Feature Transform (NDFT) is a technique designed to disentangle and separate nuisance features from important features. 
Wu \ea collected shooting condition information as metadata, which defines each environment as a "domain" and allows for classification specific to each domain~~\cite{Wu_2019_ICCV}. 
Adversarial training during backpropagation is used by introducing negative values into the backbone network, ensuring the model does not overly respond to domain-specific features but instead leverages domain-invariant features for object detection. 
Lee \ea proposed an improved version of NDFT, called A-NDFT, by introducing feature replay and a slow learner strategy to address the learning speed limitations observed in NDFT, significantly enhancing training speed while maintaining comparable performance~\cite{Lee_2021_CVPR_Workshops}. 
Jin \ea proposed the Fine-Grained Feature Disentanglement (FGFD) module, which uses the one-stage detector YOLOv5~\cite{yolov5} as the backbone network to differentiate between domain-specific and domain-invariant feature maps~\cite{rs16091626}.

There has been a significant surge in interest around vision-language representation learning tasks, largely due to training on large-scale image-text pairs, which leads to notable generalization performance. This advantage has been effectively harnessed to deliver remarkable results in image classification. Specifically, active research is underway in domains that demand high generalization capabilities, such as few-shot learning, zero-shot learning, and open-vocabulary learning.
Zhou \ea proposed the Context Optimization (CoOp) approach, which addresses concerns about the time-consuming nature of using manually crafted prompts for labeling~\cite{zhou2022coop}, similar to CLIP~\cite{radford2021learning}. To address this issue, they replaced manual prompts with learnable prompts, reducing labeling time while maintaining performance comparable to CLIP.
Building on this concept, Zhou \ea introduced Conditional Context Optimization (CoCoOP), where they incorporated a subnetwork called Meta-Net to conditionally pass information extracted from the image encoder to the learnable prompt, leading to high performance in vision-language tasks~\cite{zhou2022cocoop}.

In the field of domain-generalized object detection, a noticeable trend has emerged toward incorporating large-scale vision-language models. 
Vidit \ea introduced CLIP-the-Gap, which leverages data augmentation in the feature space through text prompts. This approach enhances generalization across diverse environments for models trained on single domains with various known domain labels~\cite{vidit2023clip}. 
Liu \ea built on this concept with LGNet, a pioneering work utilizing large-scale vision-language models for aerial image object detection~\cite{liu2024shooting}.
Their approach employs a two-step training process: First, text prompt embeddings are fine-tuned using manual prompts, and then the object detection model is trained with the fine-tuned CLIP model. 
The manual prompts incorporate domain-specific information like drone altitude, angle, and weather. 
However, this approach has a limitation in that it confines drone-specific information to only altitude, angle, and weather, neglecting other potential conditions that might impact detection performance.

Our proposed approach diverges from the traditional reliance on manual prompts by employing a learnable prompt. 
This adjustment enables us to utilize a broader spectrum of domain-specific knowledge beyond the fixed information used in previous methods. 
Moreover, we shift from the conventional two-step training process, where text prompt embeddings are fine-tuned separately before training the object detection model. 
Instead, by updating the learnable prompt concurrently with the object detection model training, we establish a more efficient, streamlined one-stage training process.
The proposed LEAP:D approach, illustrated in Figure~\ref{fig.proposed}, streamlines the training process by adopting a learnable prompt mechanism that updates concurrently with the object detection model, enabling efficient one-step training compared to traditional two-step methods.

\section{Proposed Method}
\subsection{Problem Statement}\label{sec.problem}
Given an input image $x$ and a manual prompt $x_\text{mp}$ from the training dataset $X_\text{train}$, the visual encoder $F(\cdot)$ in CLIP converts $x$ into the visual embedding $v$, while the textual encoder $G(\cdot)$ maps $x_\text{mp}$ to the textual embedding $t_\text{mp}$.
This process is formalized by the following equations: 

\begin{equation}
	v = F(x),
\end{equation}

\begin{equation}
	t_\text{mp} = G(x_\text{mp}).
\end{equation}

\ni The input manual prompt $x_\text{mp}$ follows a specific format: "An \{altitude condition\} altitude \{view condition\} view of a \{weather condition\} day taken by a drone". 
As a result, the predicted probability can be expressed as:
\begin{equation}
	p(\hat{y}=y|I) = \frac{\exp(\text{sim}(v, t_\text{mp})/\tau)}{\sum \exp(\text{sim}(v, t_\text{mp}^i)/\tau)},
\end{equation}

\ni where the function $\text{sim}(\cdot)$ measures the similarity between the visual and textual representations and is defined as:

\begin{equation}
 \text{sim}(a, b) = \frac{a \cdot b}{|a||b|}.
\end{equation}

\ni However, this method is burdened by the labor-intensive process of crafting manual prompts and requires a subsequent two-stage training process: fine-tuning CLIP first, followed by training the object detection model.

In this paper, we propose the \underline{\textbf{LEA}}rnable \underline{\textbf{P}}rompts-based domain generalized aerial object \underline{\textbf{D}}etection approach, called LEAP:D. 
This method utilizes large-scale vision-language models with learnable prompts, removing the necessity for manually crafted prompts and facilitating a streamlined one-stage training process. 
LEAP:D aims to improve efficiency and simplify model training and deployment. 
Further details of this approach are outlined below.

\subsection{Architecture Overview}
The proposed approach builds upon the LGNet architecture~\cite{liu2024shooting}, utilizing an object detection model that incorporates Faster R-CNN~\cite{girshick2015fast} and Feature Pyramid Network (FPN)~\cite{lin2017feature}. 
Additionally, the CLIP model is used during training to filter out domain-specific features. 
With intermediate features $f$ obtained from FPN, the Feature
map Squeeze Network (FSN) aligns the dimensions of the visual embeddings $v$ with the intermediate features $f$.
The aligned vector, $f' = \text{FSN}(f)$ is trained to resembe the visual embeddings $v$, while being dissimilar to textual embeddings $t$.
In this context, the similarity between the $f'$ and $v$ reflects domain-invariant features, while the similarity to $t$, generated through domain-specific prompts, indicates domain-specific features. 
During training, the object detection model learns to preserve domain-invariant features and reduce the impact of domain-specific features, resulting in a domain-generalized object detection model. 
Consequently, the trained object detection model operates independently of CLIP and during inference and remains unaffected by variations in shooting conditions.

\subsection{Domain-specific Learnable Prompt}
Domain-specific learnable prompts $x_\textit{lp}$ is constructed as follows:

\begin{equation}
 x_{lp} = [e_1][e_2]...[e_{n-1}][e_{n}],
\end{equation}

\ni where $[e_1][e_2]...[e_{n-1}][e_{n}]$ represents the learnable prompts, and we set $n = 8$. 
The textual embedding of $x_\textit{lp}$ is then denoted as $t_{lp} = G(x_{lp})$. 
The predicted probability is computed as:

\begin{equation}
	p(\hat{y}^{'}=y|I) = \frac{\exp(\text{sim}(v, t_{lp})/\tau)}{\sum \exp(\text{sim}(v, t_{lp}^i)/\tau)},
\end{equation}

\ni where we use cross-entropy loss, denoted as $L_\text{clip}$ to fine-tune the CLIP.

We follow the LGNet approach~\cite{liu2024shooting}, utilizing both similarity and dissimilarity function, defined as:

\begin{equation}
	s = (1+\text{sim}(v, f'))/2,
\end{equation}

\begin{equation}
	ds = (1-\text{sim}(t_{lp}, f'))/2,
\end{equation}

\ni where $s$ and $ds$ represent the similarity and dissimilarity functions, respectively. 
The corresponding domain-invariant and domain-specific loss functions are:

\begin{equation}
	L_{di} = -(1-s)\log{s},
\end{equation}

\begin{equation}
	L_{ds} = -\frac{1}{N_{sc}}\sum{(1-ds)\log{ds}},
\end{equation}

\ni where $L_\text{di}$ and $L_\text{ds}$ denote domain-invariant and domain-specific losses, respectively. 
The dissimilarity $ds$ and the domain-specific loss $L_\text{ds}$ should not be confused.
The loss function for the object detection model, $L_\text{od}$, is based on the Faster R-CNN~\cite{girshick2015fast}. 
The overall loss function of the proposed method is formulated as:
\begin{equation}
	L_{total} = \lambda_{1}L_{od} + \lambda_{2}L_{lp} + \lambda_{3}L_{di} + \lambda_{4}L_{ds},
\end{equation}

\ni where $\lambda_{1},\ldots, \lambda_{4}$ denote weight coefficients, set to $\lambda_{1} = 1.0$, $\lambda_{2} = 1.0$, $\lambda_{3} = 0.5$ and $\lambda_{4} = 0.5$.

\section{Experimental Results}
\subsection{Implementation Details}
The proposed method was evaluated on the VisDrone dataset~\cite{zhu2021detection}, measuring object detection performance with \(mAP_{50}\), \(mAP_{75}\) and \(mAP_{50:95}\). 
The dataset contains 8,599 images split into train and validation sets. 
Experiments were conducted on an RTX 3090 (24G) GPU.
We used SGD~\cite{ruder2016overview} as the optimization function with a learning rate $0.01$, momentum of $0.9$, a learning rate decay of $0.0001$, and trained for $12$ epochs.

\subsection{Quantitative Results}
Table~\ref{table.sota} presents the performance comparison of the proposed method with other state-of-the-art (SOTA) models on the Visdrone validation set. LEAP:D achieved $mAP_{50}$, $mAP_{75}$ and $mAP_{50:95}$ scores of 42.1, 25.5, and 24.8, respectively, demonstrating significant performance improvements compared to other SOTA networks. Specifically, compared to the baseline network LGNet, LEAP:D achieved improvements of  2.4\%p, 0.7\%p and 1.1\%p in $mAP_{50}$, $mAP_{75}$ and $mAP_{50:95}$, respectively. Similarly, compared to Faster R-CNN, our method achieved improvements of 3.6\%p, 1.6\%p and 2.1\%p in $mAP_{50}$, $mAP_{75}$ and $mAP_{50:95}$, respectively. 

\begin{figure*}[t!]
	\centering
	\includegraphics[width=7in]{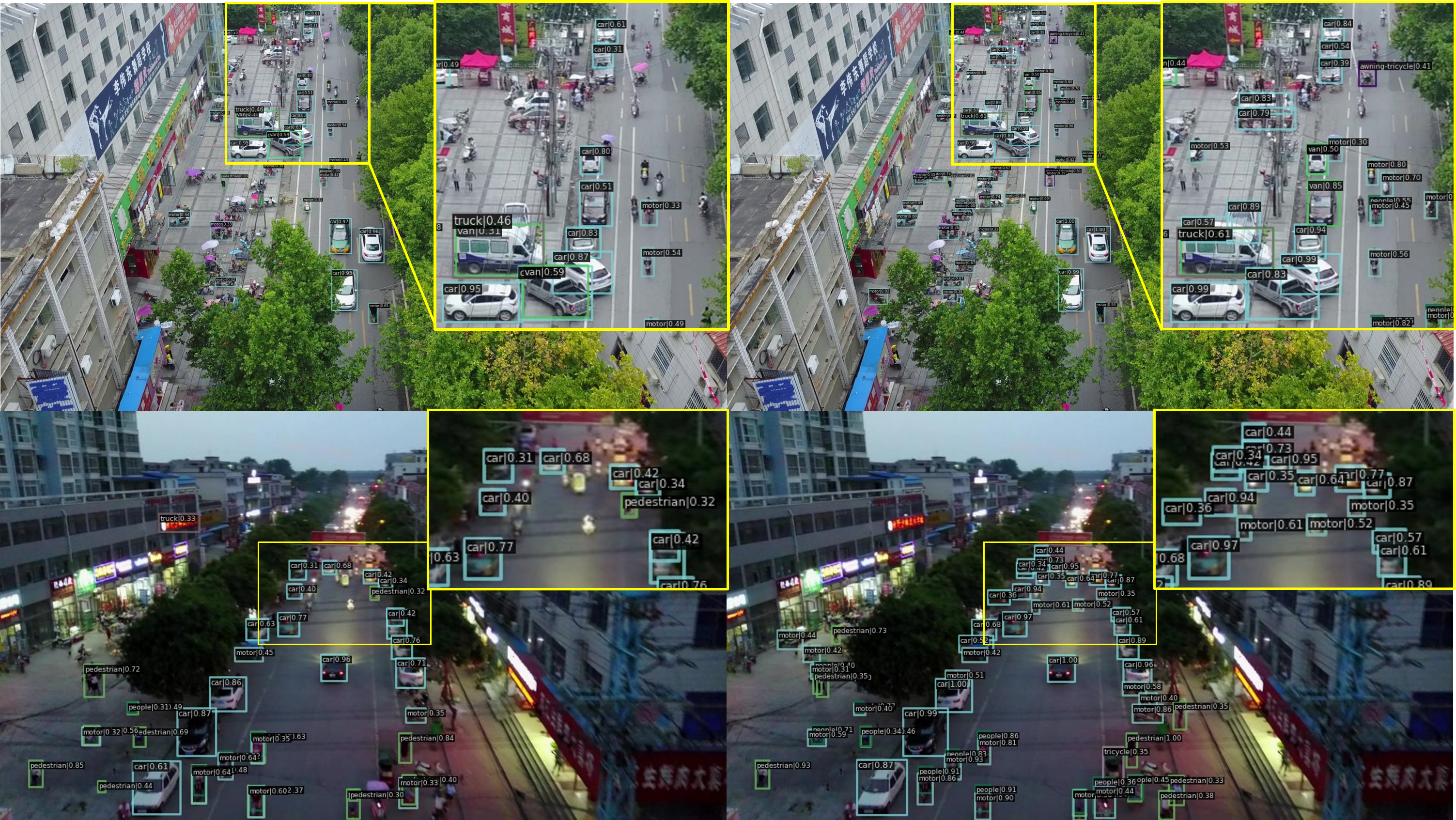}%
	\caption{The proposed method (right) and the baseline (left) are compared based on their predictions on the VisDrone dataset, as shown in the qualitative results. The \textcolor{yellow}{yellow} boxes highlight zoomed-in areas where the proposed method successfully detects objects that the baseline network misses. This observation demonstrates the superior generalization ability of the proposed method across various shooting conditions.
	}
	\label{fig.qualitative}
\end{figure*}

\begin{table}[!h]
	\caption{Quantitative Results: Performance comparison of the proposed method with other state-of-the-art (SOTA) models on the Visdrone validation set. The best scores are highlighted in bold, the second-best scores are underlined, and the \textcolor{red}{red} text indicates the performance improvement achieved by the proposed method.}
	\label{table.sota}
	\begin{center}
		{\small{
				\resizebox{8.25cm}{!}{
					\begin{tabular}{c|ccc}
						\hline
						Method     &   \(mAP_{50}\) & \(mAP_{75}\) & \(mAP_{50:95}\) \\ \hline
						FasterRCNN~\cite{girshick2015fast}  & 38.5 & 23.9 & 22.7  \\
						CascadeRCNN~\cite{cai2018cascade}  & 38.9 & 24.6 & 23.6  \\
						LGNet~\cite{liu2024shooting}  & \underline{39.7} & \underline{24.8} & \underline{23.7} \\ \hline
						LEAP:D  & \textbf{42.1}\color{red}{(+2.4)} & \textbf{25.5}\color{red}{(+0.7)} & \textbf{24.8}\color{red}{(+1.1)}         \\ \hline
						
					\end{tabular}
		}}}
	\end{center}
\end{table}

\subsection{Qualitative Results}
In Figure~\ref{fig.qualitative}, we present the qualitative results, comparing the prediction outcomes of the baseline (LGNet) and the proposed method (LEAP:D) on the VisDrone dataset. The visualizations clearly show that LEAP:D outperforms the baseline network in object detection, demonstrating superior generalization across varying shooting conditions. The baseline network relies on fixed manual prompts containing domain-specific information like altitude, view, and weather conditions. However, this fixed-domain approach may miss other relevant domain-specific details. Conversely, our proposed method uses a learning-based approach to effectively filter diverse domain-specific features and more efficiently extract domain-invariant features during training, leading to improved object detection capabilities across a wider range of environments.

\subsection{Ablation Study}
Table~\ref{table.ablation} presents the results of the ablation study, comparing the performance of the proposed method based on the number of learnable prompts. LEAP:D, with 8 learnable prompts, achieved $mAP_{50}$, $mAP_{75}$, and $mAP_{50:95}$ scores of 42.1, 25.5, and 24.8, respectively. In comparison to the manual prompts with an embedding size of 32, LEAP:D consistently outperforms the baseline network across various embedding sizes ($4\sim32$).

\begin{table}[!h]
	\caption{Ablation study showing performance comparison of the proposed method based on the number of learnable prompts. The best scores are highlighted in \textbf{bold}.
	}
	\label{table.ablation}
	\begin{center}
		{\small{
				\resizebox{8.25cm}{!}{
					\begin{tabular}{c|ccc}
						\hline
						Learnable Prompt &   \(mAP_{50}\) & \(mAP_{75}\) & \(mAP_{50:95}\) \\ \hline
						manual(32)  & 39.7 & 25.5 & 23.7   \\ \hline
						4  & 41.8 & 25.4 & 24.5 \\
						8  & \textbf{42.1} & \textbf{25.5} & \textbf{24.8} \\
						16  & 41.8 & 25.1 & 24.5 \\ 
						32  & 41.9 & 25.3 & 24.6       \\ \hline
					\end{tabular}
		}}}
	\end{center}
\end{table}

\section{Limitations and Future Works}
This study faces two main limitations. First, it removes domain-specific features from the entire scene rather than targeting them at the object level. Our future work will focus on applying learnable prompts to individual objects to reduce domain-specific features more effectively. Second, the experiments were limited to a single dataset. In the future, we plan to broaden our testing by evaluating our model on diverse datasets, including aerial imagery and images captured in a variety of environments.

\section{Conclusion}
In this study, we presented LEAP:D, a domain-generalized aerial object detection approach that leverages learnable prompts for drone imagery. 
Our comparative experiments revealed that LEAP:D outperforms baseline models and other state-of-the-art methods. 
These results highlight both the efficiency and effectiveness of our approach, emphasizing its promise as a leading solution for domain-generalized aerial object detection using large-scale vision-language models. 
By utilizing learnable prompts, our method achieves high accuracy, effectively addressing various shooting conditions and reinforcing the robustness of domain-generalized detection.


\bibliographystyle{IEEEbib}

\end{document}